\documentclass[letterpaper, 10pt, conference]{ieeeconf}      

\IEEEoverridecommandlockouts                              
\usepackage{graphicx}
\usepackage{changepage} 
\usepackage{subcaption}
\usepackage{amsmath}
\usepackage{algorithm}
\usepackage{algorithm}
\usepackage{booktabs}
\usepackage[dvipsnames,svgnames,x11names]{xcolor}

\usepackage[noend]{algpseudocode}
\usepackage{afterpage}
\usepackage{listings}
\def\BibTeX{{\rm B\kern-.05em{\sc i\kern-.025em b}\kern-.08em
    T\kern-.1667em\lower.7ex\hbox{E}\kern-.125emX}}
\usepackage{stfloats}
\usepackage{float}
\usepackage{cite}
\usepackage[hidelinks]{hyperref}

\usepackage[dvipsnames,svgnames,x11names]{xcolor}
\usepackage[acronym]{glossaries}
\usepackage{glossaries-extra}
\setabbreviationstyle[acronym]{long-short}
\setabbreviationstyle[short]{short-nolong}
\newacronym[category={short}]{pddl}{\textsc{pddl}}{The Planning Domain Definition Language}
\newacronym{gnn}{\textsc{gnn}}{graph neural network}
\newacronym[category={short}]{3dsg}{3D scene graph}{3D Scene Graph}
\newacronym[category={short}]{procthor}{\textsc{ProcTHOR}}{ProcTHOR}
\newacronym[category={short}]{alfred}{\textsc{alfred}}{Action Learning From Realistic Environments and Directives}

\begin{document}



\title{\LARGE \bf
Anticipatory Planning for Performant Long-Lived Robot in Large-Scale Home-Like Environments}

\author{Md Ridwan Hossain Talukder, Raihan Islam Arnob and Gregory J. Stein%
\thanks{Md Ridwan Hossain Talukder, Raihan Islam Arnob and Gregory J. Stein are affiliated with the CS Department, George Mason University, Fairfax, VA: {\tt\small \{mtalukd, rarnob, gjstein\}@gmu.edu}}}%

\maketitle
\thispagestyle{empty}
\pagestyle{empty}
\begin{abstract}
We consider the setting where a robot must complete a sequence of tasks in a persistent large-scale environment, given one at a time. Existing task planners often operate myopically, focusing solely on immediate goals without considering the impact of current actions on future tasks. Anticipatory planning, which reduces the joint objective of the immediate planning cost of the current task and the expected cost associated with future subsequent tasks, offers an approach for improving long-lived task planning. However, applying anticipatory planning in large-scale environments presents significant challenges due to the sheer number of assets involved, which strains the scalability of learning and planning. In this research, we introduce a model-based anticipatory task planning framework designed to scale to large-scale realistic environments. Our framework uses a \gls{gnn} in particular via a representation inspired by a \gls{3dsg} to learn the essential properties of the environment crucial to estimating the state's expected cost and a sampling-based procedure for practical large-scale anticipatory planning. Our experimental results show that our planner reduces the cost of task sequence by 5.38\% in home and 31.5\% in restaurant settings. If given time to prepare in advance using our model reduces task sequence costs by 40.6\% and 42.5\%, respectively.
\end{abstract}
\section{Introduction}
We want service robots in large-scale environments like homes, hospitals, and restaurants to perform well in the setting where a robot must complete a sequence of tasks, given one at a time. However, lacking foresight of the future tasks while executing sequences of tasks, state-of-the-art task planners \cite{garrett2020pddlstream, plaku2010sampling,toussaint2015logic,kaelbling2011hierarchical, kim2020learning, srivastava2014combined,kim2019learning,chitnis2016guided,dantam2016incremental} act myopically, focusing solely on the immediate goal without accounting for potential side effects on subsequent tasks. To address this limitation, Dhakal et al. \cite{dhakal2023anticipatory} recently presented a theoretical framework that formalizes \textit{Anticipatory Planning}—planning to reduce the joint cost of (i) completing the current task and (ii) the expected cost of future tasks—and demonstrated its potential to improve long-lived planning for small-scale, blockworld settings. Under this approach, learning is used to mitigate the computational burden of computing expected future costs, providing cost estimates that guide planning toward behaviors that avoid side effects and improve long-lived performance.

\begin{figure}
    \includegraphics[width=8.5cm]{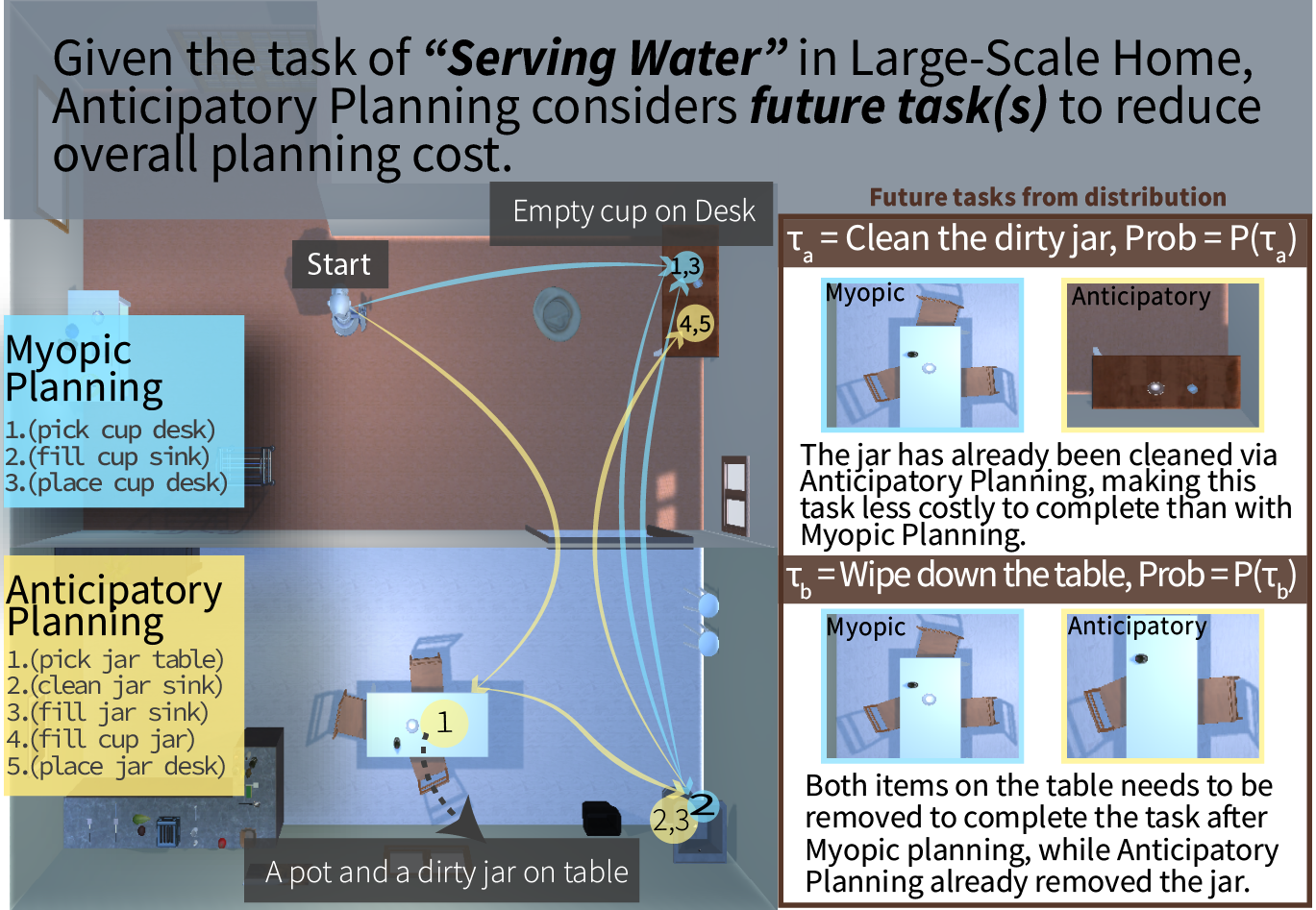}
    \centering
    \caption{\textbf{Home-Scale Anticipatory Planning: }The service robot is tasked with serving a cup of water on a desk in a large \gls{procthor} home. Next, it might be asked to `clean the jar' or `wipe the table' from the task distribution. \textbf{ Myopic Approach:} Completes the task with a lower immediate cost by using a cup to fetch and serve water. \textbf{Our Approach:} The robot cleans and fills up the jar with water and completes the task which is more costly now but reduces the cost of the future tasks thus reduces the overall cost of completing all the tasks.}
    \centering
    \label{fig:example-ap-vs-mp}
\end{figure}

However, anticipatory planning is made challenging in home-scale environments, since both the learning and planning upon which it relies struggle to scale effectively as the size and complexity of the environment grow. First, the model must effectively estimate a state's future expected cost with many assets spread across the building's multiple rooms at scale.
Second, search space for planning grows exponentially as the number of assets increases, so exploring all possible plans to accomplish a given task (e.g., Fig. \ref{fig:example-ap-vs-mp} shows two different plans that complete the task \texttt{ServeWater} at the desk) and estimating their anticipatory costs becomes impractical. Given time in advance of being assigned tasks, a robot equipped with anticipatory planning can also \emph{prepare} the environment, taking additional action to preemptively reduce expected future cost; however, this capability similarly relies on effective learning and estimation at scale. So, for anticipatory planning in a home-scale environment, we need (i) a compact representation of the world for effective learning, also required for preparation, and (ii) a planning procedure that determines suitable plans quickly and effectively handles the exponential growth of the space.

In this work, we present a learning-augmented model-based anticipatory task planning framework tailored for home-scale environments, designed to mitigate the challenges of scale for effective learning (to estimate the expected future cost) and planning (via search). Inspired by recent progress in learning and estimation in household environments, we leverage a graphical representation of the environment akin to a \gls{3dsg} \cite{Armeni_2019_ICCV} that, used to define a \gls{gnn} \cite{battaglia2018relational} estimator, is well-suited for estimating the expected cost at scale. We additionally present a search procedure for efficient anticipatory planning that considerably reduces the search space and guides our planner toward good behavior to improve long-lived task planning on a home scale. Our contributions are as follows:

\begin{enumerate}[]
\item We introduce an approach for anticipatory task planning at scale that allows robots to reduce the cost of long-horizon task sequences.
\item We developed a sampling procedure for planning to address the challenges of scaling to the home scale that reduces the search space and explicitly guides the robot to find plans with a lower expected cost.
\item We designed a compact yet semantically rich representation of the environments based on a \gls{3dsg} to define a \gls{gnn} to achieve effective learning for estimation at scale.
\item To evaluate our approach, we developed a benchmark for task planning in the home-scale \gls{procthor} \cite{procthor} with task distribution curated from the \gls{alfred} \cite{alfred} dataset and a restaurant environment of our design.
\end{enumerate}

We demonstrate that our robot significantly reduces cost over long-horizon task sequences by using our anticipatory planning framework in large-scale environments such as homes and restaurants. Compared to standard myopic planning, our approach reduces the overall cost of task sequences by 5.38\% in home and 31.5\% in restaurant environments, respectively. Additionally, if allowed to prepare the environment in advance, our planner reduces the cost by 40.6\% and 42.5\%, respectively.
\section{Related Work}
\subsection{Task Planning}
Task planning in robotics has traditionally focused on generating a set of high-level actions to complete a task within domains using domain-specific languages like \gls{pddl} \cite{fox2003pddl2, aeronautiques1998pddl} along with strategies including but not limited to search algorithms \cite{kautz1996pushing, bonet2001planning, helmert2006fast, lavalle2006planning} that use complex heuristics \cite{Hoffmann2001FFTF}. Most existing studies in this space prioritize solving problems quickly by using classical planning strategies \cite{kaelbling2011hierarchical, plaku2010sampling, srivastava2014combined}. Recently, learning-based methods have been shown to enhance planning performance, allowing for more complex plans and improving the overall quality of the plans \cite{kim2020learning, dantam2016incremental, chitnis2016guided, kim2019learning, garrett2021integrated}. However, lacking knowledge about possible future tasks, these planners do not consider the consequence of their current action on subsequent tasks and thus perform poorly when the solution of one task may affect the next task in persistent environments \cite{dhakal2023anticipatory}.

\subsection{Anticipating and Avoiding Side Effects during Planning}
Some recent works consider long-lived planning in persistent environments and show improvement in small-scale settings \cite{dhakal2023anticipatory, dhakal2024anticipatorytaskmotionplanning}. Similar research in this anticipatory planning space directly predicts future task(s) via \textsc{LLM}s \cite{arora2024anticipate} or leverages human behavior patterns \cite{patel2022proactive}  also show a promising result of anticipation while solving task planning problems. While state-of-the-art anticipatory planners have demonstrated promising performance in small problem sizes, it is not directly applicable to large-scale problems. Other works \cite{9812227, 9345470, krakovna2020avoiding}, in the space of avoiding side effects via anticipating human behavior for effective collaboration, fall outside the scope of this paper.

\subsection{Integration of Task Planning and Learning}
State-of-the-art learning-based planners have demonstrated promising performance in small-to-moderate problem sizes by integrating learning with planning\cite{chitnis2016guided, dhakal2023anticipatory, kim2020learning, kim2019learning, dantam2016incremental, shen2020learning}. However, many of these planners can struggle to scale to learn and plan for large problem instances \cite{pmlr-v164-agia22a}. Recent studies in vision-based learning show that \gls{3dsg} are useful for representing large scenes \cite{Armeni_2019_ICCV,  rosinol20203d, kim20193}. Also, studies in the area of planning show that hierarchical graph representations of the state of an environment can be used to define a \gls{gnn} to guide sample-based motion planning \cite{kim2020learning} or to estimate the relative importance of objects in a scene to accelerate the search over tasks \cite{silver2021planning}. These recent studies hold promise for estimating anticipatory cost via learning at scale.
\section{Problem Formulation}
\subsection{Preliminaries: Task Planning with PDDL}\label{pddl}
In this work, we consider task planning in home-scale environments, including rearrangement-style tasks, food serving, and cleaning tasks. We represent the tasks and the actions available to the robot using \gls{pddl} \cite{fox2003pddl2}. We define a task planning problem as a tuple $\langle \mathcal{O}, \mathcal{P}, s_0, \mathcal{A}, \tau \rangle$, where $\mathcal{O}$ represents the objects of entities (robot, containers, and interactable objects) involved, $\mathcal{P}$ denotes the set of predicates which express relations among these entities, $s_0$ is the initial state, and $\mathcal{A}$ is a set of parameterized actions such as \texttt{move}, \texttt{pick}, \texttt{place}, \texttt{clear}, \texttt{fill}, \texttt{make-coffee}, and \texttt{wash}. A task $\tau$ is defined as a list of predicates, and therefore, it defines a subset of the state space---the goal space $G_{\tau} \subseteq \mathcal{S}$---in which the task is considered complete.

The planner consumes a \gls{pddl} definition of the problem and returns the plan that solves the task. Actions like \texttt{pick}, \texttt{place}, \texttt{clear}, \texttt{make-coffee}, and \texttt{wash} have a constant cost. In contrast, \texttt{move} actions have costs proportional to their travel distance, accounting for the robot's motion from one location to another while avoiding obstacles. Motion cost for \texttt{move} actions are computed from the occupancy map via the Dijkstra algorithm. We use the notation $V_{s_g} (s_0)$ to mean `the plan cost to get from state $s_0$ to goal state $s_g$'.

\subsection{Anticipatory Planning}
The goal of anticipatory planning is to reduce the overall cost of solving a sequence of $N$ tasks in a persistent environment, given only one task at a time. The robot often has flexibility in completing its assigned task $\tau$, so the goal space $G_{\tau}$ may include many satisfying states. In long-term deployments, the environment persists between tasks, so the robot must consider how its current action to solve a task will affect future tasks and plan effectively. We use the anticipatory planning formulation of Dhakal et al. \cite{dhakal2023anticipatory}, which seeks to minimize the expected cost of an immediately available task and a single next task in the sequence assigned from an underlying task distribution $P(\tau)$:
\begin{equation}\label{eq:antplan}
\begin{split}
s^*_g & = \underset{s'_g\in G_{\tau}}{\text{argmin}} \Bigg[V_{s'_g}(s_0) + \underset{\tau'} \sum P(\tau') V_{\tau'}(s'_g)\Bigg]\\
 & = \underset{s'_g\in G_{\tau}}{\text{argmin}} \Bigg[V_{s'_g}(s_0) + V_{A.P.}(s'_g)\Bigg]
\end{split}
\end{equation}
where $V_{s'_g}(s_0)$ is the cost of moving from state $s_0$ to an intermediate goal state $s'_g$, and $V_{\tau'} (s'_g)$ is the cost of completing task $\tau'$ starting from that intermediate state $s'_g$. Here, $V_{A.P.}(s'_g)$ is the \emph{anticipatory planning cost}: the expected cost of completing a single follow-up task starting from state $s'_g$.
While the immediate cost $V_{s'_g}(s_0)$ can be computed using existing task planning solvers, the anticipatory planning cost, $V_{A.P.}(s)$, is often too challenging to compute online. So, instead, the planner estimates it via learning \cite{dhakal2023anticipatory}.

\textit{Preparation} is a capability related to anticipatory planning. Given additional time in advance of being given a task, the robot is allowed to \emph{prepare} the environment in order to reduce cost when it is later assigned a task and improve the overall performance of completing the task sequence. It can prepare the environment by rearranging it and changing the state of its entities, and it can be considered `task-free' planning. We use the formulation for preparation established by Dhakal et al. \cite{dhakal2023anticipatory} that searches over the states to get the state with the lowest expected cost:
\begin{equation}\label{eq:prep}
s^*_{p} = \underset{s \in S}{\text{argmin}} \left[ V_{A.P.} (s) \right]
\end{equation}
where $S$ is the set of all possible states in the environment.
\section{Approach: Anticipatory Planning in Home-Scale Environments}
\begin{figure}[t]
\vspace{.4em}
\includegraphics[width=8.5cm]{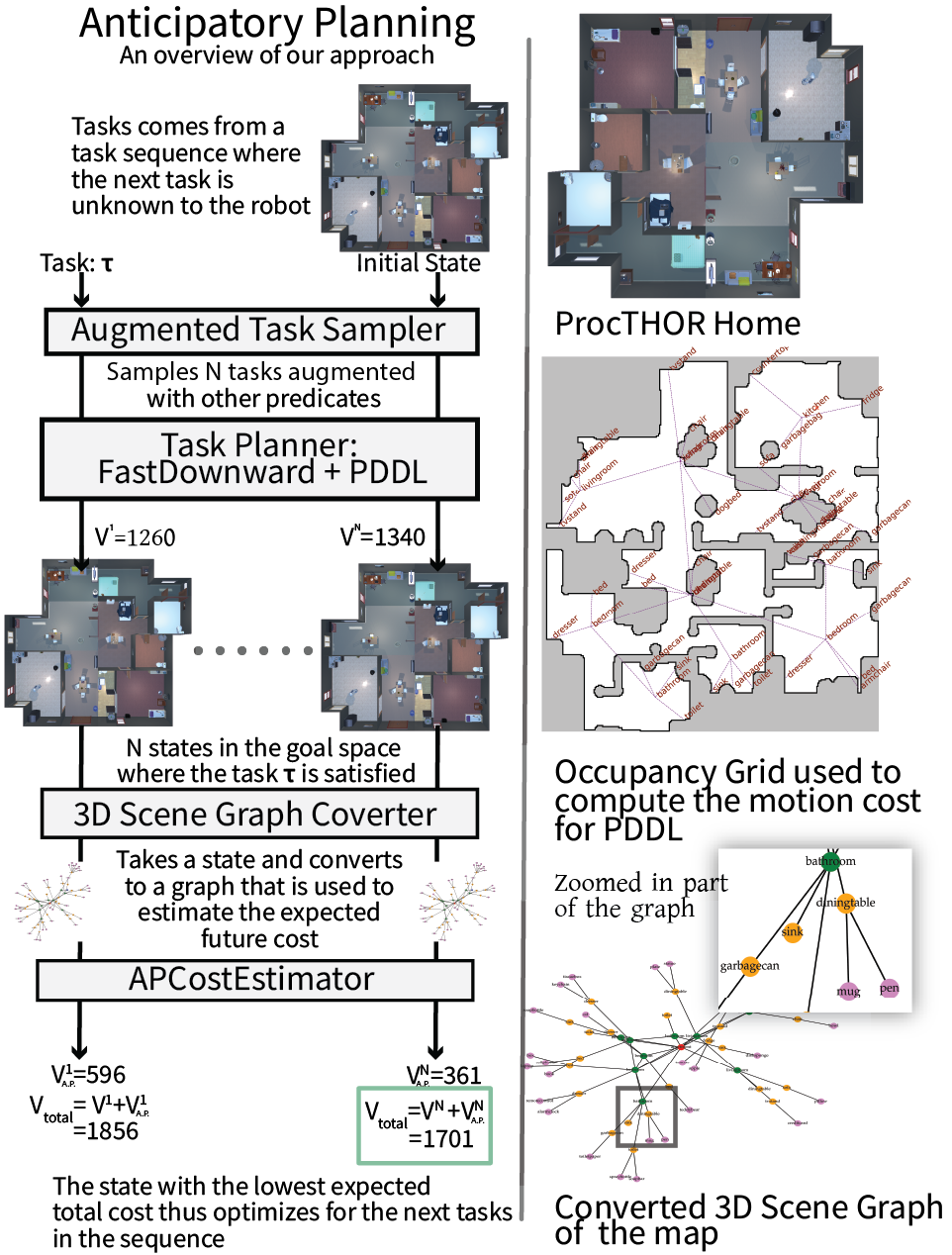}
    \centering
    \caption{\textbf{Schematic of our approach}: (Left) Shows an overview of our approach described in \ref{sec:search} and (Right) shows the maps and representations for learning described in \ref{sec:learning-3dsg}}
    \centering
    \label{fig:approach}
    \vspace{-1.5em}
\end{figure}

We present an approach for anticipatory planning at scale, planning as to minimize the sum of immediate and expected future planning costs, as defined in Eq. (\ref{eq:antplan}). Our approach involves (1) sampling plans that complete the task $\tau$ and (2) iterating over the sampled plans to get the plan that minimizes the sum of immediate plan cost and future expected cost. For example, Fig. \ref{fig:example-ap-vs-mp} shows two different plans that complete the task \texttt{ServeWater} at the desk, and the anticipatory planning approach selects the plan that minimizes the sum of immediate cost and the future expected cost. So, effective anticipatory planning in home-scale environments requires (i) a tractable procedure that effectively samples plans and (ii) an estimator that effectively estimates the future expected cost of the state. Fig. \ref{fig:approach} and the Alg. \ref{alg:cap} show an overview of our approach.

We compute the immediate plan cost via FastDownward \cite{helmert2006fast} using informed search with the \texttt{ff-astar} heuristic \cite{Hoffmann2001FFTF} and use a \gls{gnn}, in particular via a representation based on \gls{3dsg} to estimate the anticipatory cost.
We address the challenges of searching in an enormous search space in Sec. \ref{sec:search}, and Sec. \ref{sec:learning-3dsg} shows how we tackle the learning at scale issue.

\subsection{Home-Scale Anticipatory Planning via Focused Sampling}\label{sec:search}

Given a task $\tau$, multiple plans can achieve it. For example, when tasked with \texttt{ServeFruit}, the robot could use a distant clean knife to cut and serve the fruit on a plate that completes the task or take additional steps like cleaning the knife or placing a dirty bowl in the sink, anticipating future tasks. While all these plans satisfy the task, the number of plans grows exponentially with more assets, making exhaustive search intractable. To address this, we limit the sample space for plan generation.

\begin{figure}
    \includegraphics[width=8.5cm]{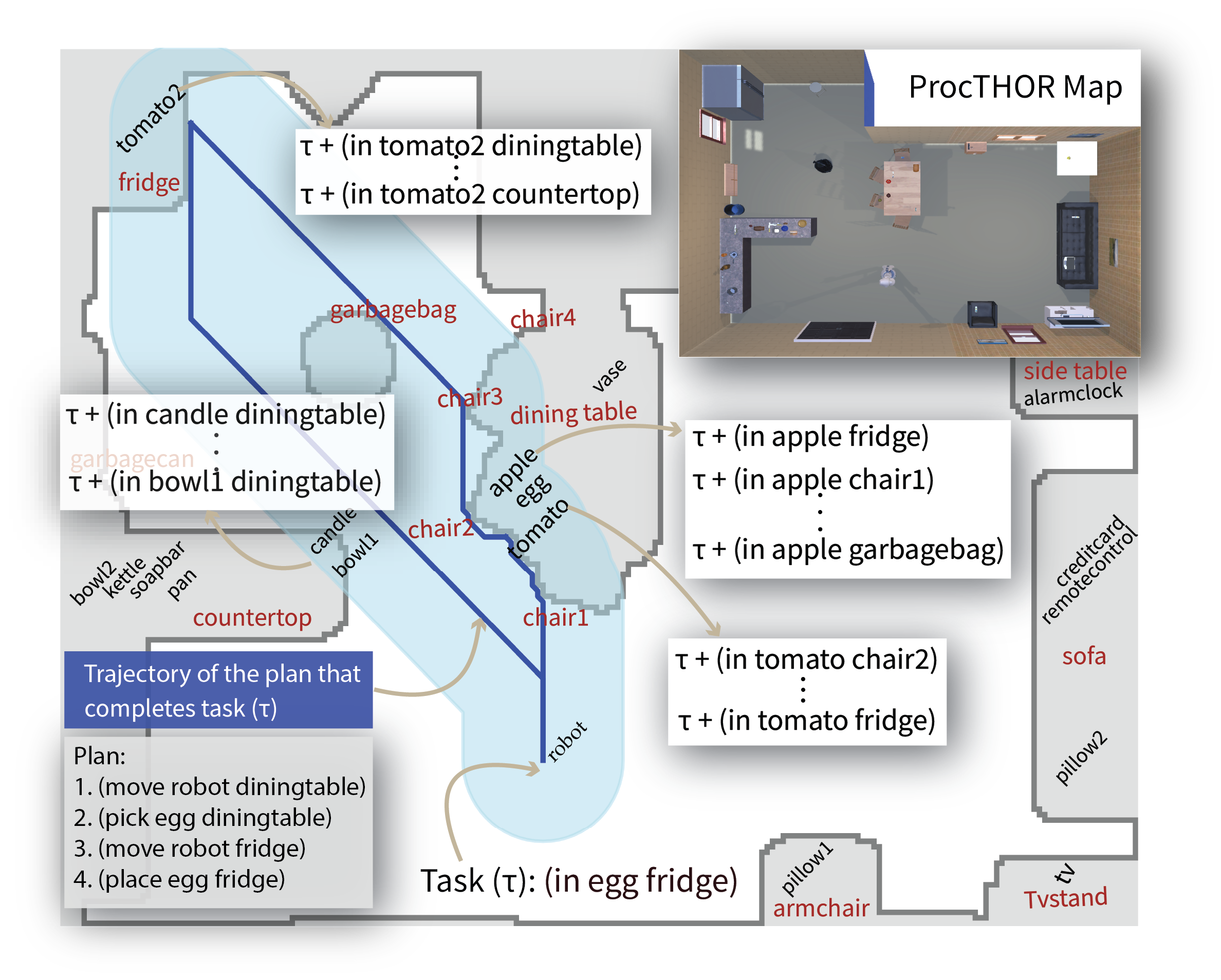}
    \centering
    \caption{\textbf{Sampling procedure:} We sample tasks for planning by augmenting the given task with other predicates only involving entities within a bounded region.}
    \centering
    \label{fig:sample}
    \vspace{-1em}
\end{figure}

To accomplish more targeted plan generation, we instead augment the robot's objective: asking the planner to solve proxy tasks that consist of the original tasks \emph{augmented with additional predicates} $p_\text{\tiny{add}}$, additional objectives the robot must accomplish during its execution. For generating the augmented tasks $\tau_\text{\tiny{aug}} \equiv \tau \cup p_\text{\tiny{add}}$, the process \textsc{SampleAugmentedTasks} uses focused sampling by including the predicates $p_\text{\tiny{add}}$ involving entities (containers and interactable objects) $o \subseteq \mathcal{O}$ that the robot encounters in its path within a bounded region while executing the given task $\tau$. Fig. \ref{fig:sample} shows a schematic of our sampling procedure in \gls{procthor} environment while completing a pick-and-place task.

\begin{figure}
    \begin{algorithm}[H]
    \caption{Anticipatory Task Planning}\label{alg:cap}
    \begin{algorithmic}[1]
    \Function {Anticipatory Task Planning}{$s_0, \tau$}
    \State $\pi^*, V_{s_g} \gets \textsc{TaskPlan}(s_0, \tau)$
    \State $s_g \gets \textsc{Tail}(\pi^*)$
    \State $\phi_g \gets \textsc{ConvertToGraph}(s_g)$
    \State $V^*_{\text{total}} \gets V_{s_g} + \textsc{APCostEstimator}(\phi_g)$
    \For {$i \in  \{1,2,...,N\}$}
    \State $\tau_{\text{aug}} \gets \textsc{SampleAugmentedTasks}(s_0, \tau, \phi_g)$
    \State $\pi, V_{s_g} \gets \textsc{TaskPlan}(s_0, \tau_{\text{aug}})$
    \State $s_g \gets \textsc{Tail}(\pi)$
    \State $\phi'_g \gets \textsc{ConvertToGraph}(s_g)$
    \State $V_\text{total} = V_{s_g} + \textsc{APCostEstimator}(\phi'_g)$
    \If {$V_\text{total} \leq V^*_\text{total} $}
    \State $V^*_\text{total}= V_\text{total}$
    \State $\pi^* = \pi$
    \EndIf
    \EndFor
    \State \Return $\pi^*$
    \EndFunction
    \end{algorithmic}
    \end{algorithm}
    \vspace{-3em}
\end{figure}

We then solve each $\tau_\text{\tiny{aug}}$ using the FastDownward that returns the plan $\pi$ and its cost $V_{s_g}$ to find the state $s_g \in G_\tau$ that minimizes the total cost. This cost includes the task completion cost $V_{s_g}$ (computed via the solver) and the anticipatory cost $V_{A.P.}(s_g)$  (via our \textsc{APCostEstimator}). To get the anticipatory cost, the method \textsc{Tail} extracts the terminal state $s_g$ from the plan $\pi$, and then \textsc{ConvertToGraph} function converts $s_g$ to corresponding graph $\phi_g$ needed for the estimator. 
Additionally, to save computation, we filter $\tau_\text{\tiny{aug}}$ that may lead to a terminal state without improving the future expected cost. To filter tasks, we systematically perturb state $s_g$ to recover a state $s'_g$ that satisfies the augmented task $\tau_\text{\tiny{aug}}$. Estimating the future expected cost of that state using \textsc{APCostEstimator}, we select only tasks that reduce costs compared to completing $\tau$ without any additional predicates.

\textit{Preparation} involves searching for a state of the environment that minimizes the expected future cost, given additional time before a task is assigned, to improve the overall performance of completing a task. We conduct this search using a simulated annealing optimization approach \cite{bertsimas1993simulated}. Starting from an initial state $s_0$, we iteratively perturb the object states $N$ times. In this process, if the anticipatory planning cost $V_{A.P.}$ of the new state, estimated by the \textsc{APCostEstimator}, is improved or satisfies a probabilistic acceptance criterion that depends on a gradually decreasing temperature factor. In that case, the new state is accepted for further perturbation. This search process continues for N iterations, after which the prepared state $s_p$ is returned.

\subsection{Estimating Anticipatory Planning Cost at Scale}\label{sec:learning-3dsg}
Inspired by recent developments in mitigating the challenges of learning from large building-like environments, we use a \gls{gnn} via a \gls{3dsg}-based representation \cite{Armeni_2019_ICCV, kim20193, 10610243, 10638815} to predict the anticipatory cost required for Anticipatory Planning. We abstract the world as sparse graph-based representations, where nodes are the entities embedded with semantic features (e.g., node’s name, type, color, state), and edges represent spatial or logical relationships among nodes. The nodes of the graph have accompanying \emph{node input features} such as the name of the entity embedded using SBERT \cite{reimers2019sentencebertsentenceembeddingsusing}, type, position on the map, and state of the entity (e.g., whether it is empty or filled with liquid). We leverage this graph representation to define the graph structure of the \gls{gnn} to estimate the expected cost of a state of the environment.


\subsection{Data Generation} \label{sec:datagen}
To train our \gls{gnn} to estimate the anticipatory cost of a state we need training data generated from similar environments.
We assume that our robot has direct access to the underlying task distribution $P(\tau)$ specific to that environment, specifying the tasks the robot may be assigned and their relative likelihood. We start from an initial state of the environment, randomly sample a set of tasks from the task distribution, and solve that using a task planning solver Sec. \ref{pddl}. We use the resulting planning cost to calculate the anticipatory planning cost using Eq. (\ref{eq:antplan}). Each datum consists of a \gls{3dsg} representation of the state with the expected cost as the label, which the learned model learns to estimate.

\section{Experiments and Results}
\begin{figure*}[t]
    \centering
    \vspace{.5em}
    \includegraphics[width=16cm]{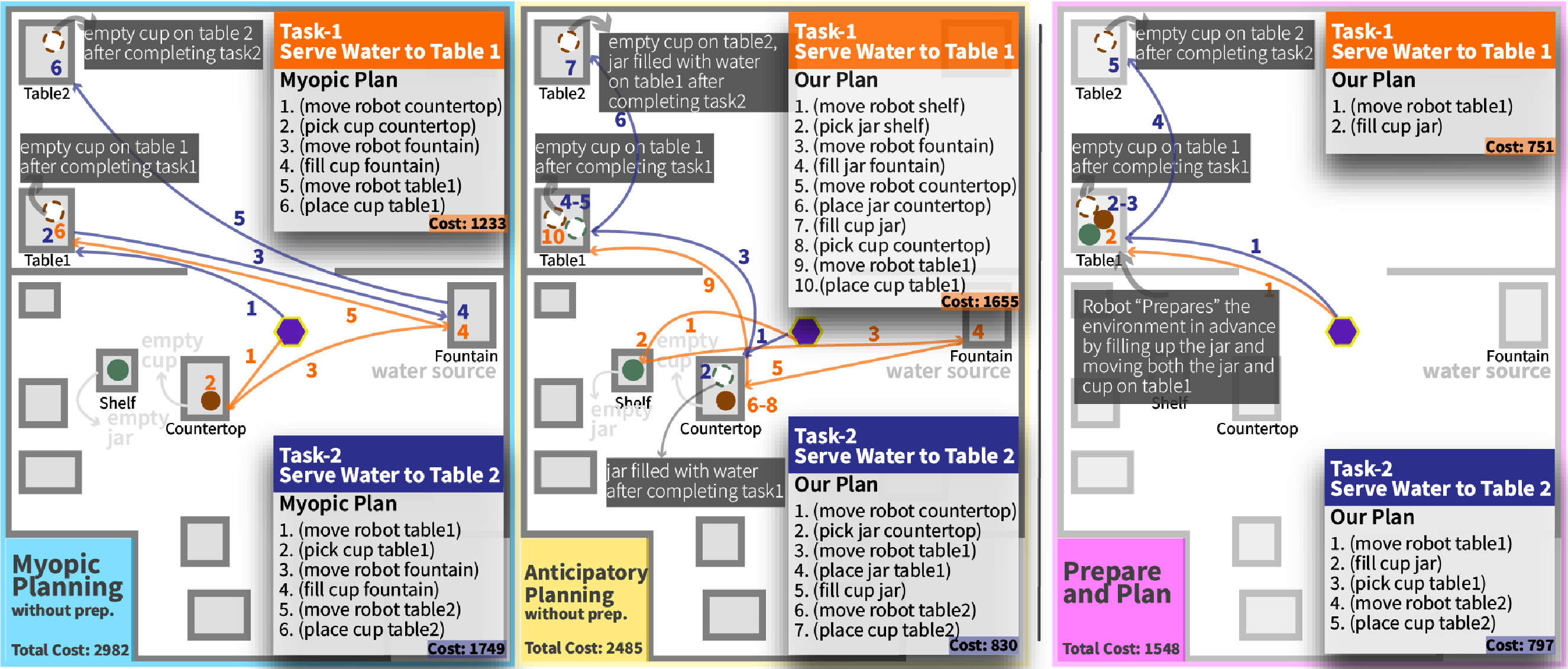}
    \caption{\textbf{Restaurant Domain Example with two tasks}: Anticipatory Planning (left) reduces the total planning cost over Myopic Planning (middle) by anticipating the future tasks. Preparation in advance (right) further reduces the planning cost.}
    \label{fig:scene-graph}
\end{figure*}
We evaluate the performance of our \textsc{APCostEstimator} for anticipatory planning and preparation in large-scale simulated \gls{procthor} and restaurant environments of our design. The task distribution for the \gls{procthor} environments are curated from the \gls{alfred} dataset, and so contains only rearrangement-style pick-and-place tasks. So, we additionally design a restaurant setting with tasks description of greater complexity and object interaction, including but not limited to serving and cleaning style tasks to evaluate the estimator's performance for anticipatory planning in large-scale restaurant domain.
We evaluate performance using four planning approaches that leverage different aspects of our learning-augmented anticipatory planning approach:
\begin{LaTeXdescription}\label{planning:approach}
  \item[Myopic Baseline] Classical planning via FastDownward.
  \item[Anticipatory Planning] Planning augmented with estimates of the anticipatory planning cost. Planning seeks to reduce the overall cost of both the immediate task and a single future task, as in Eq. (\ref{eq:antplan}).
  \item \textbf{Preparation + Myopic Baseline (Prep. + Myopic)} Myopic planning, yet the robot is first allowed to \emph{prepare} the environment using the estimator, as in Eq. (\ref{eq:prep})
  \item \textbf{Preparation + Anticipatory Planning  (Prep. +Anticipatory Planning)} Anticipatory planning, yet the robot is first allowed to \emph{prepare} the environment.
\end{LaTeXdescription}

\begin{table}[t]
    \begin{center}
    \label{table:toy-stats}
        \begin{tabular}{ccc}
        \toprule
            Evaluated Approaches & \textbf{\gls{procthor}} & \textbf{Restaurant} \\
            \hline     
            Myopic Baseline   & 260.1 & 1139.1\\ 
            Anticipatory Planning (ours)   & 246.1 & 780.6\\ 
            Prep. (ours) + Myopic Baseline   & 155.7 & 727.9\\ 
            Prep. + Anticipatory Planning (ours)   & \textbf{154.5} & \textbf{654.7}\\ 
        \bottomrule
        \end{tabular}
    \caption{Average cost per task averaged over 500 sequences (20-task per sequence in \gls{procthor} and 40-task per sequence in Restaurant)}
    \vspace{-2em}
    \end{center}
\end{table}

\subsection{Graph Neural Network Implementation}
Our estimator is a \gls{gnn} consists of four TransformerConv \cite{DBLP:journals/corr/abs-2009-03509} layers. Each layer was normalized using a batch normalization layer followed by a leaky ReLU activation function. We also combine node features within each graph in the batch into a single vector using both mean and add pooling, which helps in aggregating information from all nodes. Finally, a linear layer to produce the anticipatory cost. We use mean absolute error (MAE) as our loss function. We train our model using Adagrad optimizer and a learning rate scheduler StepLR on PyTorch~\cite{paszke2019pytorchimperativestylehighperformance} with a batch size of 8 for 10 epochs. We reduce the learning rate with a decay factor 0.5 using the fixed step size of 1000. Though the structure is the same, we have two different networks for two environments.

\subsection{Evaluating the Estimator in Home-scale Environments}  
Our homes are curated from \gls{procthor} that consist of 10K unique maps, and the task distribution for these maps comes from the \gls{alfred}, keeping only the feasible tasks for that environment. \gls{procthor} has a variety of homes, starting from 1-room apartments to 14-room home complexes, with 1,633 household objects in 108 categories. Each environment has around 50 to 200 unique pick-and-place-style tasks from \gls{alfred}. Task specifications consist of placement directives for one movable object; for example, a task could involve moving an apple to the dining table. The tasks for data generation and evaluation are taken from this task distribution. Containers are constrained to hold at most seven objects per container. We generate training data described in Sec. \ref{sec:datagen}. For homes, each node (apartment, room, containers, and objects) of the \gls{3dsg} has three features: a string of the container name embedded using SBERT, a one hot encoding of whether the node is a robot, container or movable object and its position, combined a total 776-length feature vector.

We evaluate our estimator to predict the expected cost of large \gls{procthor} homes containing numerous assets. We include experimental trials showing (i) planning beginning from a random initial configuration of the \gls{procthor} environment and also (ii) the impact of \emph{preparation}. We evaluate 20-length task sequences, drawn according to the task distribution $\tau \sim P(\tau)$, in randomly generated 500 \gls{procthor} homes for 10,000 task executions.

Our results in \gls{procthor}-\gls{alfred} setting demonstrate that our estimator effectively predicts a state's expected cost. By preparing the \gls{procthor} environments and using anticipatory planning, we achieve a substantial 40.6\% reduction in the cost of task sequences. Using anticipatory planning from the non-prepared state yields a 5.38\% improvement in cost efficiency for the task sequences compared to myopic planning. We note that preparation helps myopic and anticipatory planning reduce the total cost. However, as the tasks curated from \gls{alfred} contain only placement-style tasks, it limits the interdependency required to show the impact of anticipatory planning.

\subsection{Evaluating Anticipatory Planning in Restaurant}
To evaluate the effectiveness of our anticipatory planning approach at scale with more complex and diverse task settings, we design restaurant environments from 1K different layouts with two rooms; one is a kitchen, and another is a serving room. This setting contains containers and objects of 25 different categories. For each restaurant environment, the task distribution comes from the tasks, such as \texttt{MakeCoffee}, \texttt{ServeWater}, \texttt{ServeFruitBowl}, \texttt{ClearConatiners}, \texttt{WashObjects}, accompanied with pick-and-place type tasks. We only keep the feasible tasks for that environment; each has around 50 to 100 unique serving, cleaning, and pick-and-place type tasks. The evaluation tasks are also taken from the task distribution. Unlike the home environment, the restaurant containers have no limit on how many objects they can hold simultaneously. We generate training data described in Sec. \ref{sec:datagen}. For restaurant environments, each node (restaurant, room, containers, and objects) of the graph has four features: a string of the node's name embedded using SBERT, a one hot encoding of the type of node, its position, and a one hot encoding of 9 attributes of the assets (e.g., \texttt{isLiquid}, \texttt{isEmpty}) which combines to a 785-length feature vector.

\begin{figure}
    \vspace{.5em}
    \includegraphics[width=7.5cm]{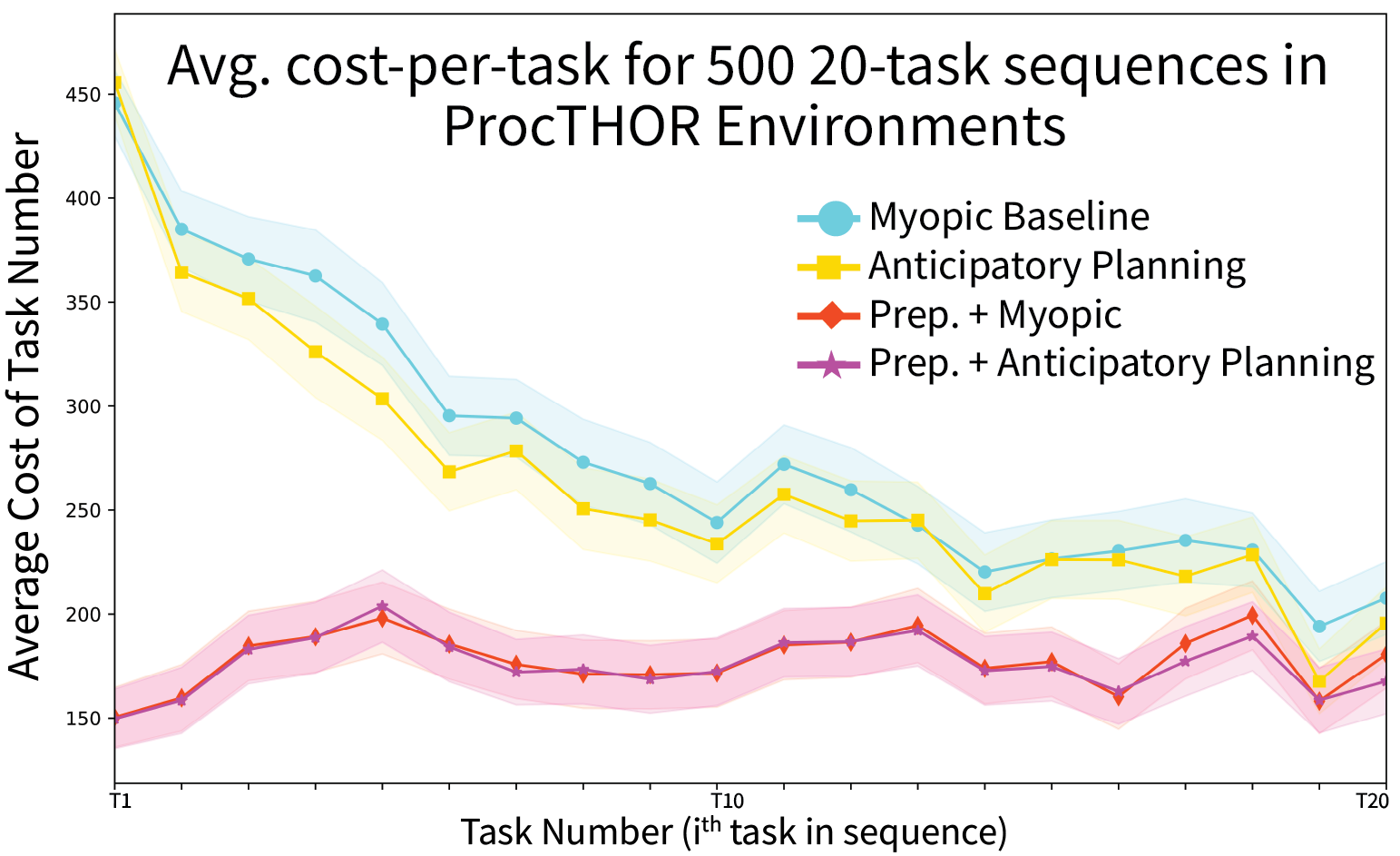}
    \centering
    \caption{Preparing the \textbf{\gls{procthor}} environments using our estimator reduces the average-cost per each task of the task-sequence compared to planning without preparation.}
    \centering
    \label{fig:res-proc}
    \vspace{-1em}
\end{figure}

For evaluation, similar to the home environment, we include experimental trials showing (i) planning beginning from a random initial configuration of the restaurant environment and also (ii) the impact of \emph{preparation}. For each restaurant environment, we evaluate 40-length task sequences drawn according to the task distribution $\tau \sim P(\tau)$ in 500 different randomly generated restaurant environments for a total of 20,000 task executions. For each trial, we evaluate performance using four planning approaches described in Sec. \ref{planning:approach} that leverage different aspects of our learning-augmented anticipatory planning approach.

\begin{figure}
    \vspace{.5em}
    \includegraphics[width=7.5cm]{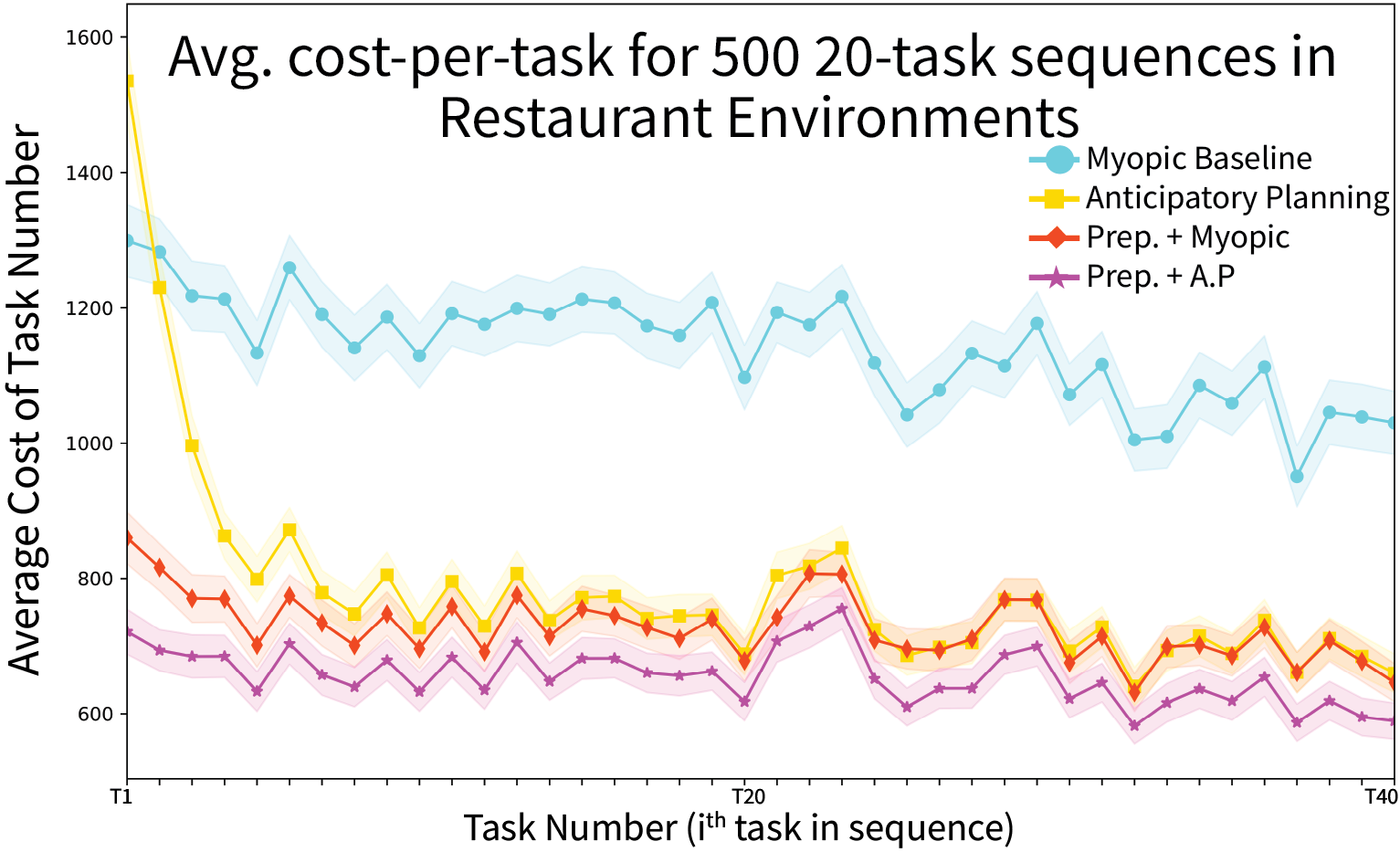}
    \centering
    \caption{Results in \textbf{Restaurant} environments show that our approach for large-scale anticipatory planning reduces the average cost per task for long-horizon task sequences.}
    \centering
    \label{fig:result-restaurant}
    \vspace{-1em}
\end{figure}

Our findings from the restaurant settings confirm that our estimator capably learns to guide the planner in effectively searching over plans for anticipatory planning and preparation. The results show that when a robot performs anticipatory planning from a non-prepared state, it can reduce the total task sequence cost by 31.5\% compared to myopic planning. Moreover, preparing the restaurant in advance leads to a 42.5\% reduction in task sequence costs. This substantial decrease highlights the impact of anticipatory planning and preparation on task planning efficiency in large-scale environments. In the Figure. \ref{fig:result-restaurant}, we also observe that when starting from an unprepared state, employing myopic planning insignificantly reduces the average cost per task over time.
In contrast, anticipatory planning gradually reduces costs and guides the robot closer to a \emph{prepared} state. We also identified the differential impact of environmental preparation on various tasks and quantified the extent of cost reduction associated with such preparation. Approximately 80\% of tasks (in both restaurant and \gls{procthor}) are less costly when the environment is \emph{prepared} in advance.

\section{Conclusion and Future Work}
We present a learning-augmented anticipatory planning approach for home-scale environments that scale to learn and plan to improve long-horizon manipulation-style task planning performance. For tractable anticipatory planning in large-scale environments, we developed a sample-based strategy that guides the planner effectively, and to learn to scale, we combined the \gls{gnn} with a \gls{3dsg}-based representation. We demonstrated that our learned model can accurately estimate the expected cost of a large state to prepare the environment in advance when no task is given and during anticipatory planning and improves the planner's performance over long-horizon tasks. 

However, we assume that the robot can always access the task distribution and that both the training and evaluation tasks come from that distribution, which may not always be true. The robot may encounter unseen tasks even though deployed in a similar environment. In future work, we may explore how robots learn the underlying task distribution and efficiently collaborate with other robots with different capabilities.
\bibliographystyle{IEEEtran}
\bibliography{references.bib}
\end{document}